\pdfoutput=1

\documentclass[11pt]{article}

\usepackage{acl}

\usepackage{times}
\usepackage{latexsym}
\usepackage{multirow}
\usepackage{rotating}
\usepackage{subcaption}
\usepackage{graphicx}
\usepackage{float}
\usepackage{amsmath}
\usepackage{multirow}
\usepackage{adjustbox}

\usepackage[T1]{fontenc}

\newenvironment{packed_itemize}{
\begin{list}{\labelitemi}{\leftmargin=1em}
\vspace{-4pt}
 \setlength{\itemsep}{1pt}
 \setlength{\parskip}{0pt}
 \setlength{\parsep}{0pt}
}{\end{list}}

\usepackage[utf8]{inputenc}

\usepackage{microtype}

\usepackage{inconsolata}

\usepackage{booktabs}

\def\Snospace~{\S{}}

\title{On-the-Fly Fusion of Large Language Models and Machine Translation }

\author{Hieu Hoang \quad Huda Khayrallah \quad Marcin Junczys-Dowmunt \\
        [2mm]  Microsoft, 1 Microsoft Way, Redmond, WA 98052, USA \\
        \texttt{\{hihoan,hkhayrallah,marcinjd\}@microsoft.com}
        }
        
\begin{document}

\setlength{\abovedisplayskip}{4pt}
\setlength{\belowdisplayskip}{4pt}

\maketitle
\begin{abstract}

We propose on-the-fly ensembling of a neural machine translation (NMT) model with a large language model (LLM), prompted on the same task and input. 
Through experiments on 4 language directions
with varying data amounts, we find that a slightly weaker-at-translation LLM can improve translations of a NMT model, and such an ensemble can produce better translations than ensembling two stronger NMT models.
We demonstrate that our ensemble method can be combined  with various techniques from LLM prompting, such as in context learning and translation context.

\end{abstract}

\section{Introduction}

For many English NLP tasks, LLMs  \cite{NEURIPS2020_1457c0d6,smith2022using,chowdhery2022palm,touvron2023llama1,touvron2023llama} are the clear state-of-the-art---e.g. sentiment analysis \cite{zhang2023sentiment},  summarization \cite{zhang2023benchmarking}.
However, dedicated NMT outperforms all but the largest closed source LLMs \cite{jiao2023chatgpt} and dedicated MT is stronger in low resource settings \cite{hendy2023good,robinson2023chatgpt}.

We propose a novel integration of a LLM and dedicated NMT model via token-level fusion. This ensembling combines strengths of each model, which emerge from their differences.
LLMs are  trained on more data than NMT models, and have more parameters. While LLMs are exposed to some parallel data \cite{briakou-etal-2023-searching}, they are trained on vastly more monolingual data, which likely gives them different domain coverage and more fluency than dedicated models. NMT models are trained on the translation task. For example, \citet{jiao2023chatgpt} found ChatGPT is more likely to hallucinate but is stronger at translating the spoken domain, while dedicated  models are stronger for medical domains and  social-media-style noisy text.  LLMs  can easily be prompted with auxiliary information--- such as domain and document context---while that is more complicated for NMT.

\noindent In this work we: 
\begin{packed_itemize}
    \item propose on-the-fly ensembling of an MT model with a prompted-for-translation LLM,
    \item combine it with  domain and context prompting,
    \item demonstrate that a weaker-at-translation LLM can improve translations of a MT model,
    \item and demonstrate our method is better than  MT ensembles and ensembles with non-prompted LLMs. 
\end{packed_itemize}

\section{Method}
\label{sec:Model}
We review standard inference of encoder-decoder NMT models and decoder only LLMs and then introduce 
our proposed ensemble of the two. 
\vspace{-2pt}
\paragraph{Standard Decoding}
In  encoder-decoder NMT,
the probability of token $t$ at the $i^{th}$ time step is:
\begin{equation}
p_{\textsc{MT}}(t_i)= p_{\textsc{MT}}(t_i|t_{j<i}, S)
\end{equation}
This conditions on  source sentence $S$ as the input to the encoder  and $t_{j < i}$  as the previously generated target tokens in the MT model decoder. 

When using a decoder only LLM for translation,
the probability of token $t$ at the $i^{th}$ time step is:
\begin{equation}
p_{\textsc{LLM}}(t_i)= p_{\textsc{LLM}}(t_i|M, S, t_{j<i})
\end{equation}
The concatenation of the prompt $M$, source sentence $S$ and the previous generated targets are all decoder outputs. The LLM model is prefix-decoded through the prompt and source, and then allowed to produce the target tokens. The LLM prompt $M$ can also include additional content.

\vspace{-2pt}
\paragraph{Proposed Ensemble}

When combining the two for our ensemble, we have:  
\begin{equation}
p_{\text{\scriptsize ensemble}}(t_i) = \lambda p_{\text{MT}}(t_i) + (1 - \lambda) p_{\text{LLM}}(t_i) 
\label{equ:Model}
\end{equation}
In the ensemble, $p_{\text{MT}}$ and $p_{\text{LLM}}$  condition on the tokens previously generated by the ensemble. 
$p_{\text{LLM}}$ still conditions on the prompt, which can be used to infuse the model with auxiliary information (e.g. domain or context).  
$p_{\text{\scriptsize ensemble}}$ reduces to the LLM when $\lambda = 0$ and to the MT model when $\lambda = 1$.

\begin{table}[ht]
\begin{center}
 \addtolength{\tabcolsep}{-3pt}
\begin{adjustbox}{max width=\linewidth}{
\begin{tabular}{l|rrrr} \toprule
              & German & Russian & Turkish & Hausa \\ \midrule
Train      & 290.4m & 38m & 49.5m & 600k \\ \midrule
Valid    & 1000/1002  & 1000/1002   & 3007    & 2000 \\
              & WMT21  &  WMT21      & newstest2017 & newsdev2021 \\ \midrule
Test      & 1984/2037   & 2016/2037   & 3000/3602       & 4456/4459 \\
            & WMT22         &  WMT22       & newstest2018 & newstest2021 \\ \bottomrule
TED-100      & -   & 1132   & -       & - \\ \midrule
ParaPat      & 2000   & 2000   & -       & - \\ \midrule
CTXPro  & 2000   & 2000   & -       & - \\ \bottomrule
\end{tabular}
}\end{adjustbox}
\end{center}
\caption{Size of datasets used in this work. All numbers are in sentences, except for CTXPro, which is reported in paragraphs. For the validation and testsets that are different in each translation direction, numbers listed are for $*$$\rightarrow$en/en$\leftarrow$$*$.}
\label{tab:corpora}
\end{table}

\section{Experimental Setup}
\label{sec:exp}
We aim to understand how our proposed method performs in high and low resource settings with strong  models, and design our experimental setup accordingly.

The parallel and monolingual training data for German and Russian is from the WMT22 \cite{kocmi-etal-2022-findings}\footnote{\url{https://www.statmt.org/wmt22/}} shared task. The Hausa data is from WMT21 \cite{akhbardeh-etal-2021-findings}.\footnote{\url{https://www.statmt.org/wmt21/}} The Turkish evaluation data was based on WMT18 \cite{bojar-etal-2018-findings}\footnote{\url{https://www.statmt.org/wmt18/}} and training data also includes  additional data from OPUS \cite{TIEDEMANN12.463}, excluding Paracrawl \cite{banon-etal-2020-paracrawl}, since such noisy data \cite{khayrallah-koehn-2018-impact} would require filtering \cite{koehn-etal-2018-findings,koehn-etal-2019-findings,koehn-etal-2020-findings,sloto-etal-2023-findings}.

As  domain-specific test sets we use TED-100 \cite{salesky2021mtedx} and ParaPat \cite{soares-etal-2020-parapat}. We also use TED-100 and CTXPro \cite{wicks-post-2023-identifying} for document-level experiments.\footnote{For CTXPro, we select a random sample of 2000 paragraphs for our experiments to reduce compute usage.}

\autoref{tab:corpora} summarizes the parallel training, evalution and test data and \autoref{tab:monolingual.corpora} in the Appendix summarizes the monolingual data.

We use back translation \cite{sennrich-etal-2016-improving} (with a 1:1 ratio of parallel to synthetic data) for all language pairs. 
We train Transformer `big' models for German, Russian and Turkish, and `base' for Hausa \cite{NIPS2017_3f5ee243} in Marian NMT \cite{junczys-dowmunt-etal-2018-marian}.\footnote{We convert models from Marian to Hugging Face format.} 
We use Llama2 \cite{touvron2023llama} with 7 and 13 billion parameters as LLMs. 
The LLama2 32k token SentencePiece  model \cite{DBLP:conf/emnlp/KudoR18} is used for source and target MT tokenization.\footnote{The target side vocabs must match between the LLM and MT model to be able to ensemble; the source could potentially be different. Preliminary experiments, however, found it better to use the same vocab and be able to tie the embeddings.}

The optimal mixing ratio is learnt using grid search $\lambda \in \{0, 0.1, ... 1\}$ on the validation set. We use this same value of $\lambda$ in domain specific experiments in \autoref{sec:analysis}; we do not re-sweep for each domain. Final results are reported on the test sets, translation quality is measured using \textsc{Comet-22} \cite{rei-etal-2022-comet}.
We use greedy search for decoding. 
See  \autoref{app:exp} for additional experimental details, including prompts.

\section{Results}
\label{sec:Results}

\begin{table*}[t]
\centering
 \addtolength{\tabcolsep}{-.5pt}
\begin{tabular}{l|c|cc|ccc|cc|ccc} \toprule
            &  \multicolumn{1}{c|} {MT}          
            & \multicolumn{2}{c|}{LLM 7B}
            & \multicolumn{3}{c|}{ Ensemble w/ LLM 7B} 
            & \multicolumn{2}{c|}{LLM 13B}  
            & \multicolumn{3}{c}{Ensemble w/ LLM 13B} \\ 
            
            & 
            & 0-shot & 5-shot 
            & $\lambda$ & 0-shot & 5-shot 
            & 0-shot & 5-shot 
            & $\lambda$ & 0-shot 
            & 5-shot \\ \midrule
            
\multicolumn{1}{r|} {column:}         & \multicolumn{1}{c|}{1}  
            & \multicolumn{1}{c}{2}   & \multicolumn{1}{c|}{3}  
            & \multicolumn{1}{c}{4}   & \multicolumn{1}{c}{5} & \multicolumn{1}{c|}{6} 
            & \multicolumn{1}{c}{7} & \multicolumn{1}{c|}{8}
            & \multicolumn{1}{c}{9} & \multicolumn{1}{c}{10} 
            & \multicolumn{1}{c}{11} \\ \midrule
            
de-en       & 83.5
            & 82.6    & 82.8
            & 0.7 & 83.9    & 83.9 & 82.6   & 83.4 
            & 0.7   & \textbf{84.1} 
            & 84.0  \\
en-de       & 85.4
            & 79.4    & 79.8
            & 0.7 & 85.5    & 85.5 & 63.4   & 82.4 
            & 0.8       & \textbf{85.6} 
            & \textbf{85.6} \\ \midrule

ru-en       & 82.8
            & 82.5    & 82.5
            & 0.5 & 84.0    & 84.1 
            & 81.4    & 83.4 
            & 0.5       & 84.2 
            & \textbf{84.5} \\
en-ru       & 83.1
            & 80.4    & 81.1
            & 0.5 & 83.9    & \textbf{84.2} 
            & 36.4    & 81.1 
            & 0.8       & 83.6 
            & 83.7 \\ \midrule

tr-en       & 87.2
            & 75.2    & 75.7
            & 0.8 & 87.2    & 87.2 
            & 78.9    & - 
            & 0.8       & \textbf{87.3}
            & \textbf{87.3} \\
en-tr       & 89.4
            & 57.8    & 58.2
            & 1.0 & 89.4    & 89.4 
            & 40.3    & 69.4 
            & 0.9     & 89.4
            & \textbf{89.5} \\ \midrule

ha-en       & \textbf{60.1}
            & 47.0    & 49.3
            & 0.3 & 54.7    & 54.7 
            & 46.9    & 49.7 
            & 0.3       & 54.7 
            & 54.5 \\
en-ha       & \textbf{63.1}
            & 33.1    & 37.6
            & 1.0 & \textbf{63.1}    & \textbf{63.1} 
            & 38.2    & 35.7 
            & 1.0       & \textbf{63.1} 
            & \textbf{63.1} \\ \bottomrule
\end{tabular}

  \caption{\textsc{Comet-22} on  WMT test sets. Ensembling MT \& LLM can improve scores in high resource settings where the LLM's \textsc{Comet} is somewhat worse than the MT. $\lambda$ is the mixing rate; higher $\lambda$ puts more emphasis on  MT.}
  \vspace{-5pt}
  \label{tab:results}
\end{table*}

\autoref{tab:results} shows the translation quality of the ensemble  using the 7 billion parameters LLM (col.~1-6). When both  models are of reasonable quality (de-en, ru-en, en-ru), ensembling (col.~5) results in  better quality than either  alone (col.~1 \& 2). 

In all cases, the LLM quality is worse than the MT model but ensembling with it improves most language directions. For de-en, the MT model is 0.9 \textsc{Comet}  stronger than the LLM. The ensemble still improves over the MT model by 0.6 \textsc{Comet}.

The improvement is minor for en-de, where the LLM was 21.9 points worse than  MT. The LLM translation quality for Turkish in both direction is poor while the MT is good so the ensembles are essentially reduced to the MT model. Both  models are bad for Hausa and the ensembles are unusable.
\autoref{sec:app_lambda} shows the effect of $\lambda$ on translation quality.

\textbf{In-context learning:} 
\autoref{tab:results} (col.~3) shows 5-shot learning tends to improve LLM quality but has little affect on the ensemble (col.~6). 

\textbf{Larger LLM:} \citet{xu2023paradigm} found that Llama-13B suffers from off-target issues, degrading translation out-of-English compared to the 7B model. We confirm their results---\autoref{tab:results} (col.~2 vs 7)---and also reproduce their solution of using 5-shot learning, which can recover and sometime improve LLM quality (col.~8). However, ensembling with the MT model does not require the use of in context learning (col.~10 vs 11). In general, the larger language model is better for the ensemble as de-en, en-de and ru-en all improve. It should also be noted that the
MT model adds, at most, 3\% to the number of parameters of the 7B LLM allowing the ensemble to outperform the nearly 2x bigger 13B LLM.

Ensembles for Turkish and Hausa are still not worthwhile due to the poor LLM quality in these lower resource settings. 
We use the 7B model in all analysis for the remainder of this work. 

\section{Analysis}
\label{sec:analysis}

\subsection{MT Model Ensembling}
 Given the compute resource required to use LLMs (not to mention train them), we compare the results of the MT + LLM ensemble to ensembling two MT models. 
We create ensembles for German and Russian language pairs consisting of two MT models.\footnote{The models differ only in the random seed.}
As~\autoref{tab:2marian} shows, using the LLM gives stronger translation quality in all cases except en-de, which is where the LLM underperforms the MT model by 6 \textsc{Comet} points. In all the other situations, \textit{it is better to ensemble the MT model with an LLM, even though the $2^{nd}$ MT model has higher translation quality than the LLM by 0.5 to 2.8 \textsc{Comet}.}
This suggests that when selecting models for an ensemble, simply choosing the two highest quality models is insufficient. Instead, ensembling takes advantage of the training diversity in the models to improve quality. 

\subsection{Mixing Ratio Interpretation}

The learnt mixing ratio, $\lambda$, can be loosely interpreted as a relative utility of the underlying models. For ensembles with German and Russian, $\lambda$ of 0.7 and 0.5 for the 7B LLM ensemble reflect the nearly equal contribution of both models. Due to off-target issue described above, the 13B LLM are poor at translating into German and Russian so its contribution to the ensemble is reduced.

For Turkish and Hausa, the LLM offer negligible benefit so most weight is given to the MT model. The mixing ratio space for Hausa-English is flat (see ~\autoref{fig:7b}(g)) as both underlying models are equality poor so no interpretation should be attached to the results.

\begin{table}
\centering
\addtolength{\tabcolsep}{-2pt}
\begin{adjustbox}{max width=\linewidth}{
\begin{tabular}{l|cc|c|c|c}
\toprule
        & \multicolumn{2}{c|}{MT}           & \multicolumn{1}{c|}{LLM}   & \multicolumn{1}{c|}{MT+LLM}   &\multicolumn{1}{c}{MT+MT} \\ \midrule
de-en   & 83.5&83.7   & 82.6 & \textbf{83.9}   & 83.8 \\
en-de   & 85.4&85.4   & 79.4 & 85.5  & \textbf{85.7} \\
ru-en   & 82.8&83.0   & 82.5 & \textbf{84.0}   & 83.1 \\
en-ru   & 83.1&83.2   & 80.4 & \textbf{83.9}   & 83.4 \\ \bottomrule
\end{tabular}
}\end{adjustbox}
  \caption{\textsc{Comet-22} score for two MT replicas, the LLM, the MT \& LLM ensemble, and the ensemble of the two MT models. The ensembling of the LLM with the MT model has the highest \textsc{Comet} score in all but one language pair, even though both the MT models have higher translation quality than the LLM.}
  \label{tab:2marian}
\end{table}

\subsection{Domain Prompting}
\label{sec:Prompting}

\begin{table}[t]
\centering
\addtolength{\tabcolsep}{-2pt}
\begin{adjustbox}{max width=\linewidth}{
\begin{tabular}{cl|r|rr|rr} \toprule
          &  & \multicolumn{1}{c|}{MT}        
            & \multicolumn{2}{c|}{LLM}
            & \multicolumn{2}{c}{Ensemble} \\ 
   &   prompt:      &  \multicolumn{1}{c|}{none} & \multicolumn{1}{c}{general} & \multicolumn{1}{c|}{+domain} & \multicolumn{1}{c}{general} & \multicolumn{1}{c}{+domain} \\ \midrule
 \rule{0pt}{1ex}
\multirow{1}{*}{\rotatebox[origin=c]{90}{\small TED}} &ru-en & 77.3 
            & 78.0    & 78.5    
            & 78.7    & \textbf{78.9}
            \\ [.5ex] \midrule
\multirow{4}{*}{\rotatebox[origin=c]{90}{ParaPat} } &de-en & 79.7 
            & 77.1    & 78.0    
            & \textbf{80.0}    & \textbf{80.0} \\ 
&en-de & 79.1 
            & 73.8    & 73.8    
            & \textbf{79.2}    & \textbf{79.2} \\ 

&ru-en & 72.2 
            & 74.5    & 73.9    
            & \textbf{75.1}    & 75.0 \\ 
&en-ru & 78.5 
            & 73.7    & 73.4    
            & \textbf{79.0}    & 78.7 \\ \bottomrule
            
\end{tabular}
}\end{adjustbox}

\caption{Prompting with domain can improve \textsc{Comet-22} for the LLM, but is  less effective for the ensemble.}
  \vspace{-.3cm}
\label{tab:domain.prompt}
\end{table}

The flexibility of LLM prompting can be used to add more descriptive task-specific instructions to improve quality \cite{pmlr-v202-zhang23m}. Here, we prompt for domain (TED talks and patents).

\autoref{tab:domain.prompt} shows that  additional domain information does not guarantee better LLM quality. For the TED-100 test set, ensembling has a 0.2 \textsc{Comet} improvement from an 0.5 LLM increase. Ensembling with or without the domain information in the prompt outperforms either the MT and LLM models alone. For TED, the LLM is stronger than the dedicated MT models, in contrast to our main results.\footnote{In this work, we used the $\lambda$ set on the general validation set. Re-sweeping for each specific domain could lead to improved performance.} While our dedicated MT models were not trained translation for this specific domain, the LLM likely exposed to monolingual data in this domain.  This highlights the complementary strengths of each paradigm---the ensemble leverages both.

\subsection{Document Context}
\label{sec:Context-aware}

For document or discourse input---such as TED talks---where the previous translated sentences are often relevant to the sentence to be translated, it may be better to provide the previous sentences and their translation. This contrasts with few-shot prompting where sentences pairs are high quality translations written by humans but are drawn from the validation so may not be relevant to the sentence at hand. Using sentence pairs from the same document should allow the LLM to enforce consistency across sentences and allow it to better translate phenomena that requires document-level context such as pronoun disambiguation.

\begin{figure}[]
\centering
\includegraphics[width=\linewidth]{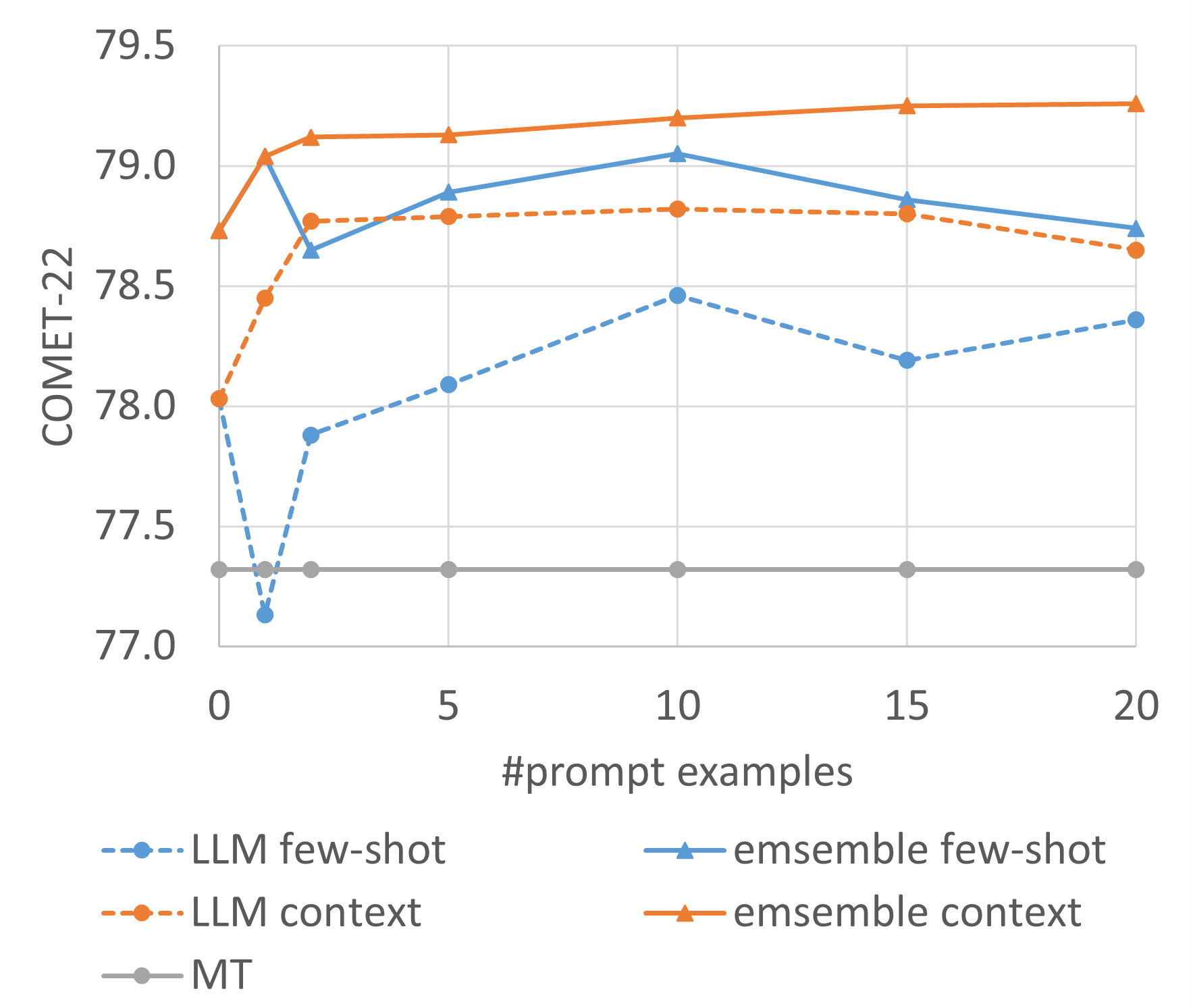}
\caption{TED-100 translation quality for various number of prompt examples (for few short learning or past context). Prompting with context outperforms few shot prompting, and it performs best when ensembled. }
\label{fig:ted100.context}
\end{figure}

\autoref{fig:ted100.context} shows the \textsc{Comet-22} score against the number of sentence pairs in the prompt on the TED-100 test set. Prompting the LLM with document context outperforms few-shot prompting and the ensemble with context (solid orange line) to outperform all variants of ensembling and LLMs with context or few-shots, as well as the MT model. \textit{Conditioning on the model's own previous outputs from the same document context outperforms few-shot prompting with the human references of less related sentences.}

Prior work found that document level-specific evaluation is required  to evaluate document level phenomena \cite{laubli-etal-2018-machine,toral-etal-2018-attaining,vernikos-etal-2022-embarrassingly}.
To this end, we use CTXPro \cite{wicks-post-2023-identifying}, a  specialized test suite which evaluates the translation accuracy of targeted words, given the document context. 

\begin{table}[t]
\centering
\addtolength{\tabcolsep}{-1pt}
\begin{adjustbox}{max width=\linewidth}{
\begin{tabular}{cl|r|rr|rr} \toprule
           & &\multicolumn{1}{c|} {MT}        
            & \multicolumn{2}{c|}{LLM}
            & \multicolumn{2}{c}{Ensemble} \\ 
          & \multicolumn{1}{r|}{context:} & \multicolumn{1}{c|}{none}& \multicolumn{1}{c}{none} & \multicolumn{1}{c|}{10 sent} & \multicolumn{1}{c}{none} & \multicolumn{1}{c}{10 sent} \\ \midrule
\multirow{3}{*}{\rotatebox[origin=c]{90}{en-de}} 
&auxiliary   & 4.5\%     & 7.2\%     & \textbf{28.0\%}  & 6.2\%   & 13.7\% \\
&formality   & 41.9\%    & 38.2\%    & 37.6\%  & 42.7\%  & \textbf{43.8\%} \\
&gender      & 44.6\%    & 38.5\%    & 39.0\%  & \textbf{45.8\%}  & 45.5\% \\ \midrule

\multirow{4}{*}{\rotatebox[origin=c]{90}{en-ru}} 
&auxiliary   & 2.6\%     & 2.3\%     & \textbf{24.6\%}  & 2.6\%   & 20.9\% \\
&formality   & 42.5\%    & 42.6\%    & 46.4\%  & 46.4\%  & \textbf{50.0\%} \\
&gender      & 27.4\%    & 31.9\%    & 36.4\%  & 31.6\%  & \textbf{37.6\%} \\
&inflection  & 28.9\%    & 22.6\%    & 25.6\%  & 29.2\%  & \textbf{31.4\%} \\ \hline

\end{tabular}
}\end{adjustbox}
\caption{CTXPro accuracy. The ensembled models with context perform  particularly well in to Russian. }
\label{tab:CTXPro.accuracy}
\end{table}

\autoref{tab:CTXPro.accuracy} shows accuracy for  various phenomena on en-ru and en-de. Adding context improves accuracy in all-but-one test set. Ensembling with context has the highest accuracy in 4 of 7 models. See \autoref{sec:app_CTXPro.comet} for \textsc{Comet} on this data; the ensemble is always best. So, when balancing \textsc{Comet} and CTXPro accuracy, the ensemble is best.

\subsection{Unprompted LM Ensembling }
\citet{yee-etal-2019-simple} and  \citet{petrick2023documentlevel} improve translation
by ensembling with a smaller-scale language model \emph{without} a task-specific prompt.

\begin{figure}[]
  \begin{subfigure}{.8\linewidth}
    \includegraphics[width=\textwidth]{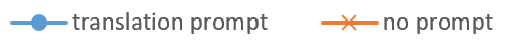}
  \end{subfigure}
  \centering
    \begin{subfigure}{\linewidth}
    \centering
    \includegraphics[width=\linewidth]{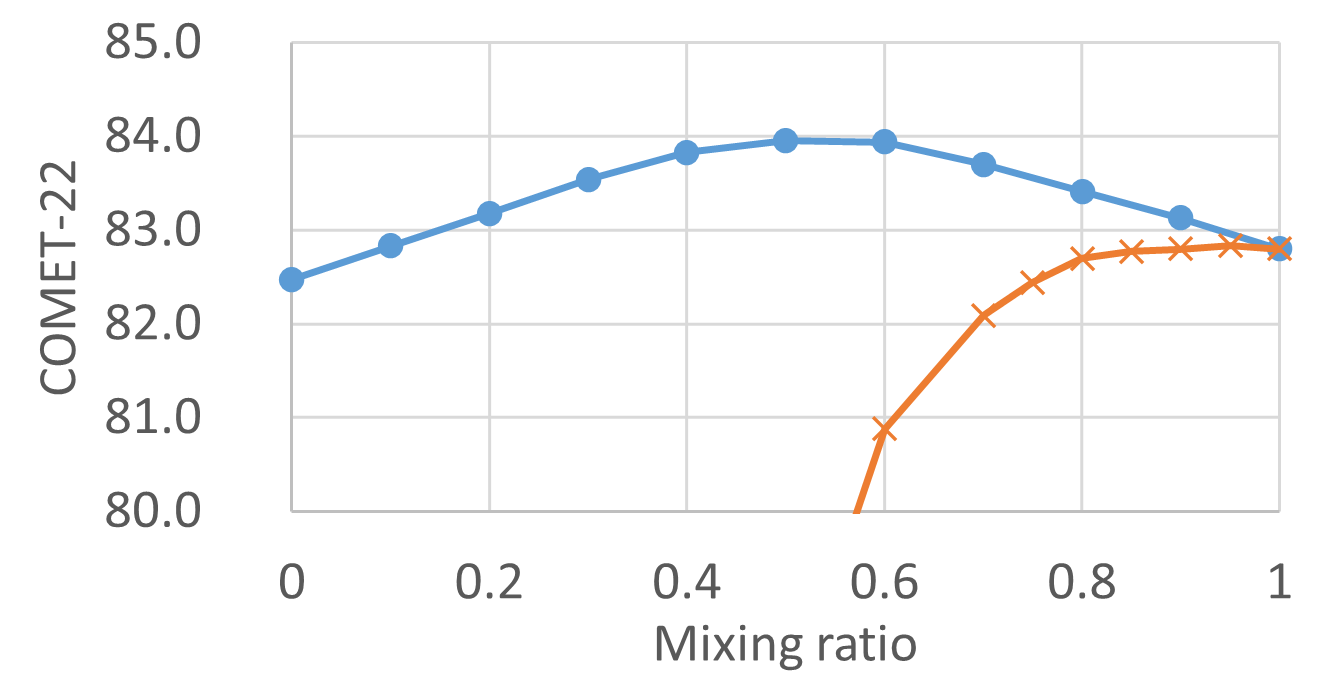}
    \vspace{-.3cm}
  \end{subfigure}%

  \caption{Using an LLM with a translation prompt and without any prompting (ru-en). Unprompted the ensemble is strictly worse than the MT baseline (mixing ratio $\lambda=1$).}
  \label{fig:no-prompt}
  \vspace{-10pt}
\end{figure}

We test this by ensembling the MT model with an unprompted LLM. \autoref{fig:no-prompt} shows that this causes quality to drop precipitously. 
The divergence from prior work may be due to differences in the base models; for example, \citet{petrick2023documentlevel} used an MT model trained on small amount of data, and \citet{yee-etal-2019-simple} trained their own LM. In our scenario with a strong MT model and a general purpose LLM, we do not see any benefit from using the LLM purely as a language model. 

\section{Related work}
\label{sec:prior_work}
\paragraph{LLMs for MT:}
Pretrained LLMs can be prompted directly for translation \cite{NEURIPS2020_1457c0d6,vilar-etal-2023-prompting,hendy2023good,robinson2023chatgpt,pmlr-v202-zhang23m,agrawal-etal-2023-context},  or fine-tuned for MT \cite{li2023eliciting,chen2023improving,moslem-etal-2023-adaptive,zeng2023tim,xu2023paradigm,yang2023bigtranslate}. Our approach is complimentary---we leverage prompting and in-context learning. We could also ensemble with a fine-tuned model. Since  we perform inference-time combination of the LLM, we do not have the same training-compute burden as fine-tuning.

Much work has explored integrating language models and NMT in various ways \cite{gulcehre2015using,GULCEHRE2017137,stahlberg-etal-2018-simple,yee-etal-2019-simple,petrick2023documentlevel}, mostly by purely conditioning a language model on the target tokens; in contrast we focus on pretrained LLMs and \textit{prompt} the LLM to produce translations.

\paragraph{Ensembling:}

Diverse inputs can be combined to create stronger ensembles \cite{Hansen1990IEEE,Dietterich2000Ensemble}. 
Various model-combination methods have been used in MT.

 System combination of outputs was used for statistical machine translation (SMT)
 \cite{Bangalore2001consensus,Heafield-marathon,freitag-etal-2014-jane}, 
and averaging model weights \cite{junczys-dowmunt-etal-2016-amu} or 
ensembling \cite{chung-etal-2016-character} are used for NMT. We build upon the latter. 
\citet{jiang-etal-2023-llm} propose a separate model to combine outputs from LLMs. We ensemble  on-the-fly. \citet{ormazabal2023comblm} ensemble two LLM from the same family where the smaller LM was finetuned for MT. We create a hybrid ensemble of two distinct architectures and training regimes.

Knowledge distillation \cite{Buciluǎ2006Model,hinton2015distilling} inspired methods can be a  way to incorporate diverse models during training \cite{dakwale-monz-2017-fine,khayrallah-etal-2018-regularized,khayrallah-etal-2020-simulated}, as opposed to during inference.
 \citet{jiang-etal-2023-llm} introduce a separate model that combines outputs from LLMs. We ensemble  on-the-fly.

 There are various methods proposed for improving translation quality by combining the adequacy and fluency  advantages of SMT and NMT \cite{devlin-etal-2014-fast,mi-etal-2016-vocabulary,junczys-dowmunt-etal-2016-amu,stahlberg-etal-2017-neural,Wang2017aaai,khayrallah-etal-2017-neural,ding-etal-2017-jhu,Zhang2021IEEE}. We combine the  strengths of NMT and LLMs.

\section{Conclusion}
We propose an on-the-fly ensembling of a dedicated MT model with an LLM, conditioned on the source and prompted for translation. 
We demonstrate that an LLM can improve translation quality of a NMT model even if the LLM is weaker at translation, provided the LLM is good enough.  
We prompt the LLM to imbue the sentence-based MT model with document-level ability, improving on sentence-level and context-focused metrics.
We find that ensembling with an LLM performs better than ensembling two MT models, even if each MT model is stronger than the LLM. 

While this work focuses on MT, the same techniques can be explored for other tasks, and may be especially useful for situations where the LLM and task-specific model have different properties and strengths.

\section{Limitations}

While we covered four languages to and from English, this is nowhere near enough to be a representative sample of languages and translation directions that would be of interest to others. We used Llama2; there are closed-access models that may be stronger at translation (e.g. GPT-4) but API access is insufficient for this method. As open-source new models are released, this method can be applied to them as well.

We used a single value of $\lambda$---which was set on the general domain validation set---for all experiments. We did not re-sweep for each domain. While this is a more general scenario that applies when test-time domain is unknown, results might be improved for focused domains by tuning $\lambda$ on domain-specific validation sets. 

In \autoref{sec:analysis}, we explore different domains (TED talks, subtitles, and patents), and use \textsc{Comet-22} as a metric.  \citet{zouhar2024finetuned} recently demonstrated that neural fine-tuned metrics, such as \textsc{Comet} are not robust to domain shift, but noted that \textsc{Comet} still had the highest overall correlation with human judgements  in their domain of study.

\bibliography{anthology,custom,mt2}

\clearpage

\appendix
\section*{Appendix}
\label{sec:appendix}
\section{Experimental Details}
\label{app:exp}

\subsection{Hyperparameters}
\label{sec:app_hyperparams}
For German, Russian and Turkish, Transformer `big' models were trained (6 layer encoder-decoder, 1024 embedding dimensions, 4096 feed-forward dimensions, 16 heads) \cite{NIPS2017_3f5ee243}. The base Transformer architecture was used for Hausa (6 layer encoder-decoder, 512 embedding dimensions, 2048 feed-forward dimensions, 8 heads). We use weight tying \cite{press-wolf-2017-using}. We train models using Marian NMT \cite{junczys-dowmunt-etal-2018-marian}.\footnote{\url{https://marian-nmt.github.io/}} We convert MT models from Marian to Hugging Face format, to allow for inference with Llama2 \cite{touvron2023llama} in the Hugging Face library \cite{wolf-etal-2020-transformers}. 

\subsection{Monolingual  Data }
\label{sec:app_data}

\begin{table*}[ht]
\begin{center}
\begin{tabular}{l|r|r|r|r|r|r|r|r} \toprule
            & \multicolumn{2}{c|}{German}
            & \multicolumn{2}{c|}{Russian}
            & \multicolumn{2}{c|}{Turkish}
            & \multicolumn{2}{c}{Hausa} \\ \midrule
            & en & de
            & en & ru
            & en & tr
            & en & ha \\ \midrule
news-commentary-v18 & 0.9m & 0.5m 
            & 0.9m & 0.5m    
            & & 
            & & \\
europarl-v10 & 2.3m & 2.1m
            & & 
            & & 
            & &  \\
news (all)  & 257.2m & 468.9m
            & 257.2m & 142.7m
            & & 
            & &  \\
news.2016   & & 
            & & 
            & 18.2m & 1.7m 
            & & \\
news.2017   & & 
            & & 
            & 26.8m & 3.0m 
            & & \\
news.2018   & & 
            & & 
            & &  
            & 18.1m & \\
news.2019   & & 
            & & 
            & &  
            & 33.6m & \\
news.2020   & & 
            & & 
            & &  
            & 41.4m & \\
CommonCrawl  & & 
            & & 
            & & 511.2m 
            & & 8.5m \\ \bottomrule
\end{tabular}
\end{center}
\caption{Monolingual Datasets.}
\label{tab:monolingual.corpora}
\end{table*}

\subsection{Prompting}
\label{sec:app_prompts}
 \autoref{fig:baseline.prompt}, \autoref{fig:domain.prompt}, \autoref{fig:fewshot.prompt}, and \autoref{fig:context.prompt} describe the various prompts we use.
\begin{figure*}[p]
  Translate the following sentence from \{src-language\} to \{tgt-language\}:
  
  \{src-language\}: \{src\}
  
  \{tgt-language\}:
  \caption{Baseline translation prompt.}
  \label{fig:baseline.prompt}
\end{figure*}

\begin{figure*}[p]
  Translate the following sentence from \{src-language\} to \{tgt-language\} in a \{style\} style:
  
  \{src-language\}: \{src\}
  
  \{tgt-language\}:
  \caption{Translation prompt with domain.}
  \label{fig:domain.prompt}
\end{figure*}

\begin{figure*}[p]
  Translate the following sentence from \{src-language\} to \{tgt-language\}:

  \{src-language\}: \{src-1\}
  
  \{tgt-language\}: \{tgt-1\}

  ...

  \{src-language\}: \{src-n\}
  
  \{tgt-language\}: \{tgt-n\}

  \{src-language\}: \{src\}
  
  \{tgt-language\}:
  \caption{n-shot translation prompt.}
  \label{fig:fewshot.prompt}
\end{figure*}

\begin{figure*}[p]
  Translate the following sentence from \{src-language\} to \{tgt-language\}:

  \{src-language\}: \{previous-src-n\}
  
  \{tgt-language\}: \{previous-translation-n\}

  ...

  \{src-language\}: \{previous-src\}
  
  \{tgt-language\}: \{previous-translation\}

  \{src-language\}: \{src\}
  
  \{tgt-language\}:
  \caption{Context-aware translation prompt.}
  \label{fig:context.prompt}
\end{figure*}

\subsection{$\lambda$}
\label{sec:app_lambda} 
\begin{figure*}[p]
\centering
  \begin{subfigure}{\textwidth}
    \centering
    \includegraphics[width=.2\linewidth]{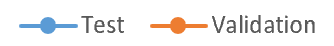}
  \end{subfigure}

  \begin{subfigure}{.25\textwidth}
    \centering
    \includegraphics[height=2.5cm,width=4.0cm]{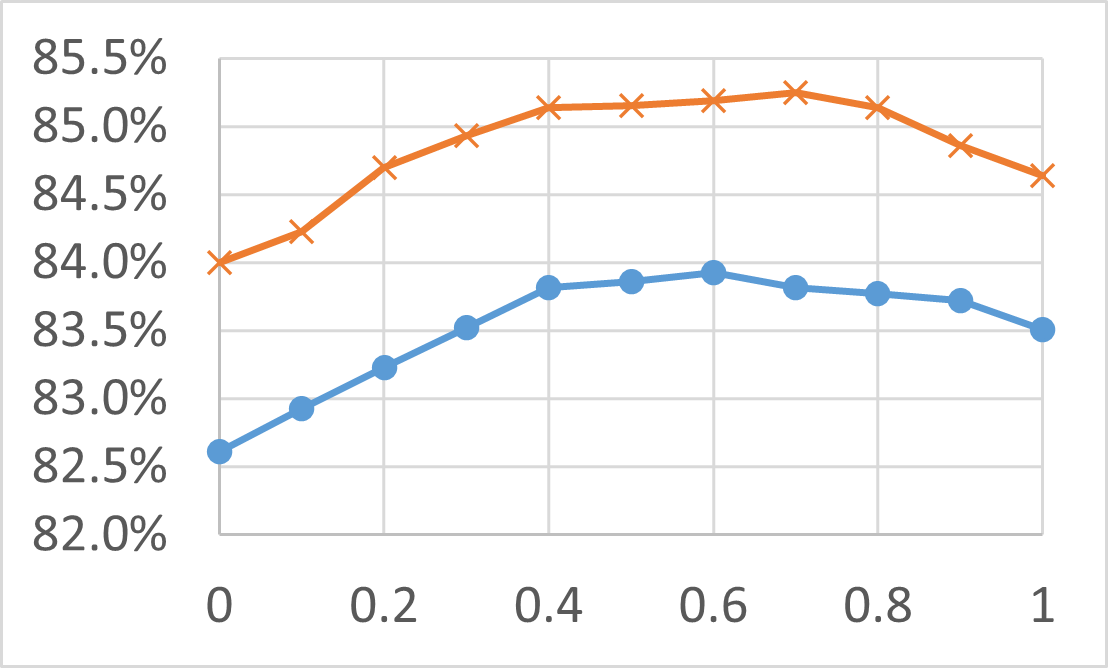}
    \vspace{-.3cm}
    \caption{de-en}
    \label{fig:sfig1a}
  \end{subfigure}%
  \begin{subfigure}{.25\textwidth}
    \centering
    \includegraphics[height=2.5cm,width=4.0cm]{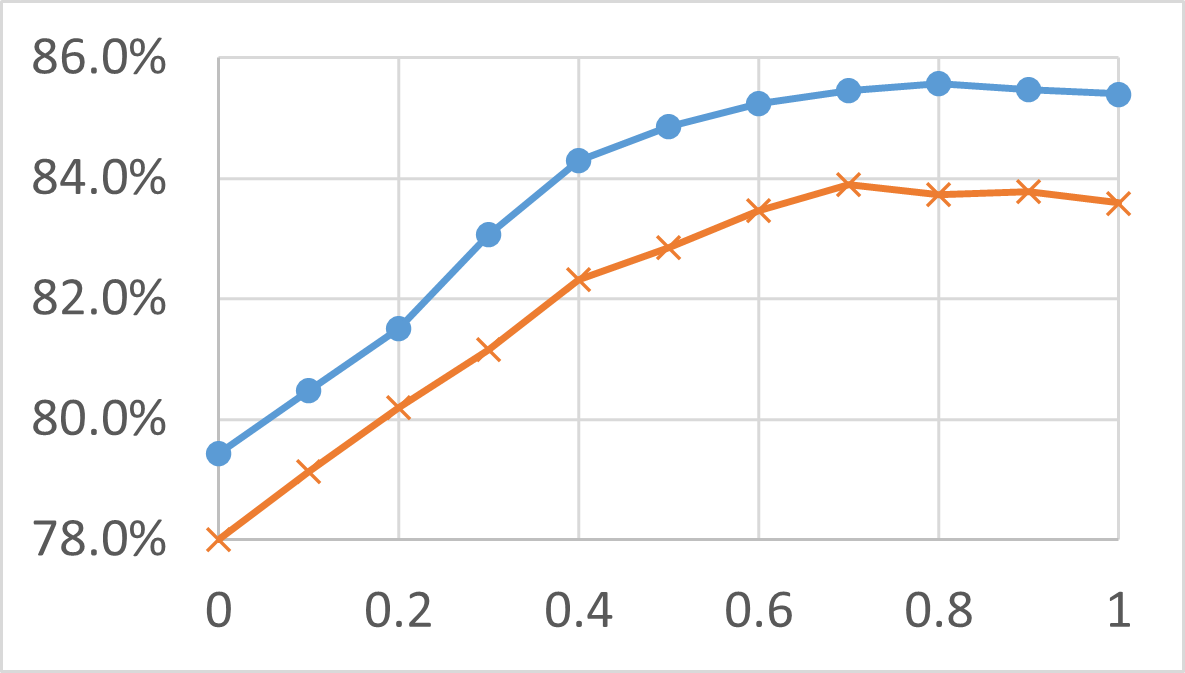}
    \vspace{-.3cm}
    \caption{en-de}
  \end{subfigure}%
  \begin{subfigure}{.25\textwidth}
    \centering
    \includegraphics[height=2.5cm,width=4.0cm]{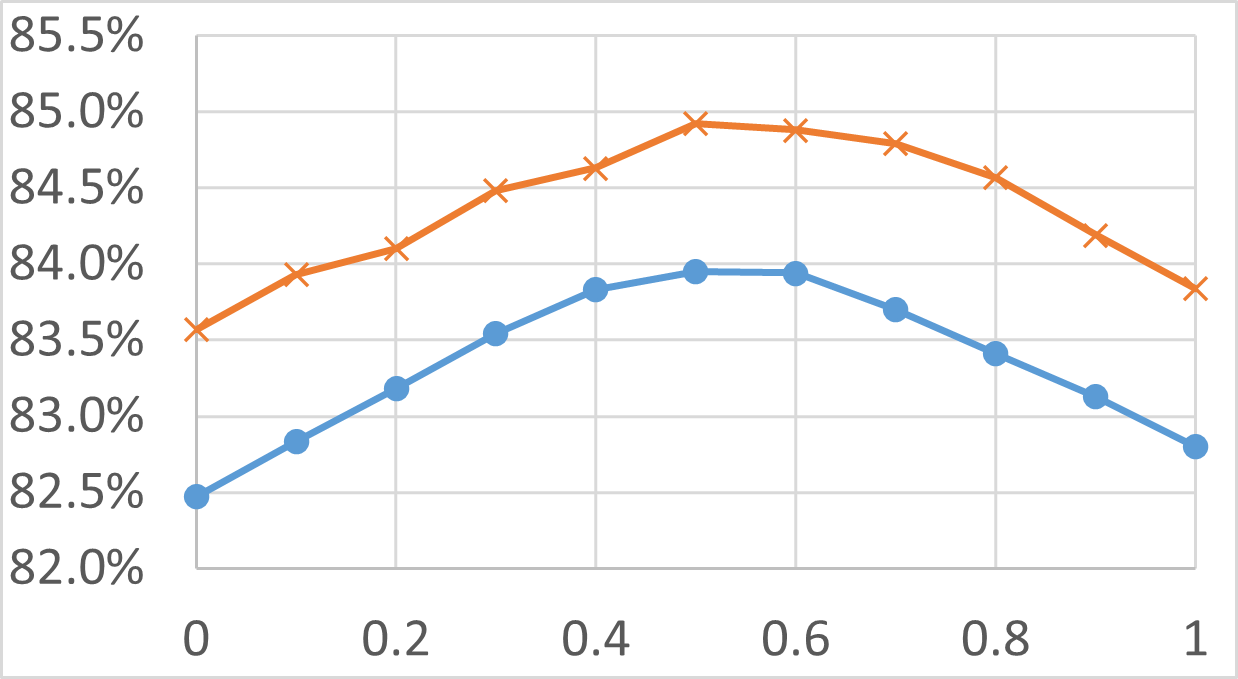}
    \vspace{-.3cm}
    \caption{ru-en}
  \end{subfigure}%
  \begin{subfigure}{.25\textwidth}
    \centering
    \includegraphics[height=2.5cm,width=4.0cm]{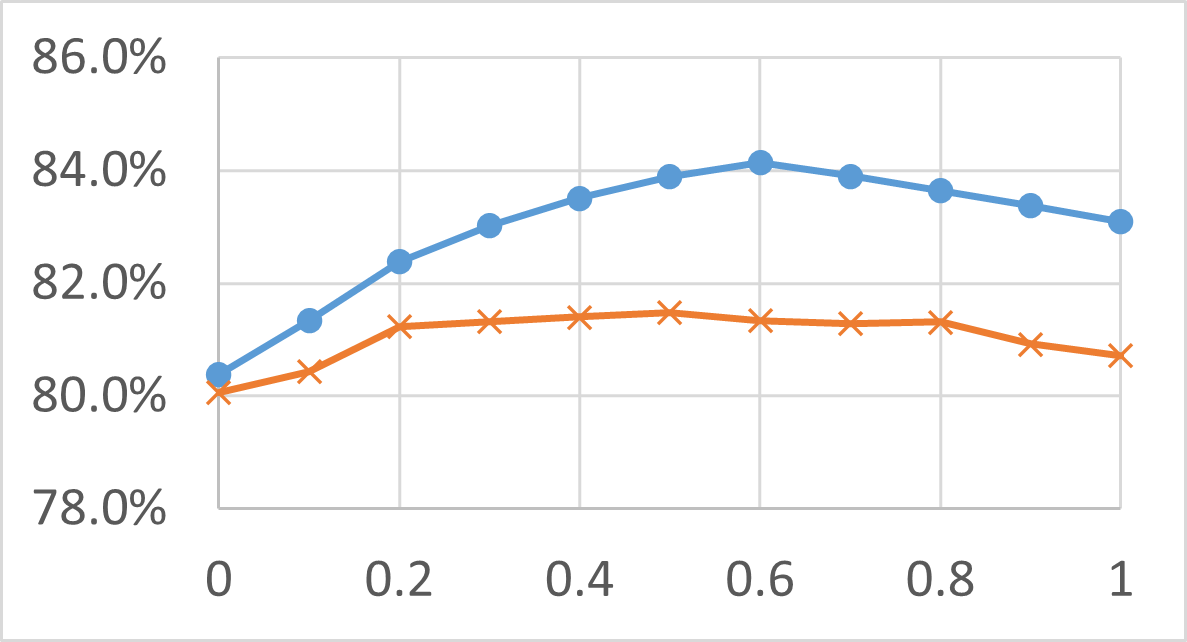}
    \vspace{-.3cm}
    \caption{en-ru}
  \end{subfigure}%

  \begin{subfigure}{.25\textwidth}
    \centering
    \includegraphics[height=2.5cm,width=4.0cm]{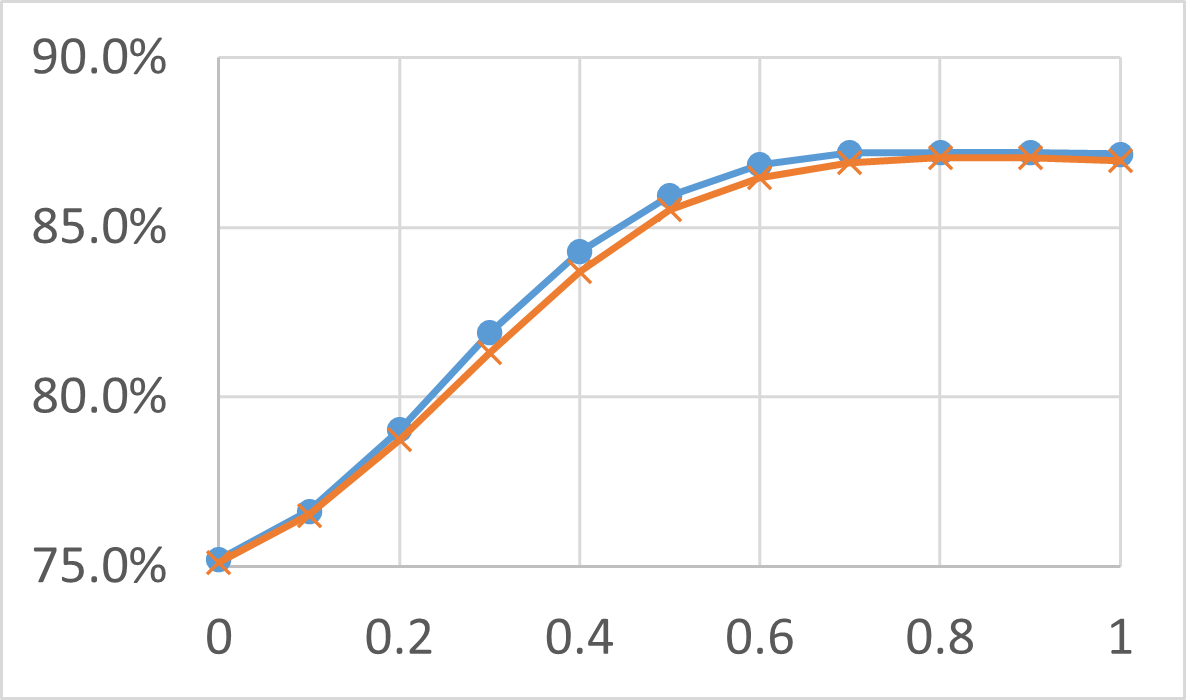}
    \vspace{-.3cm}
    \caption{tr-en}
  \end{subfigure}%
  \begin{subfigure}{.25\textwidth}
    \centering
    \includegraphics[height=2.5cm,width=4.0cm]{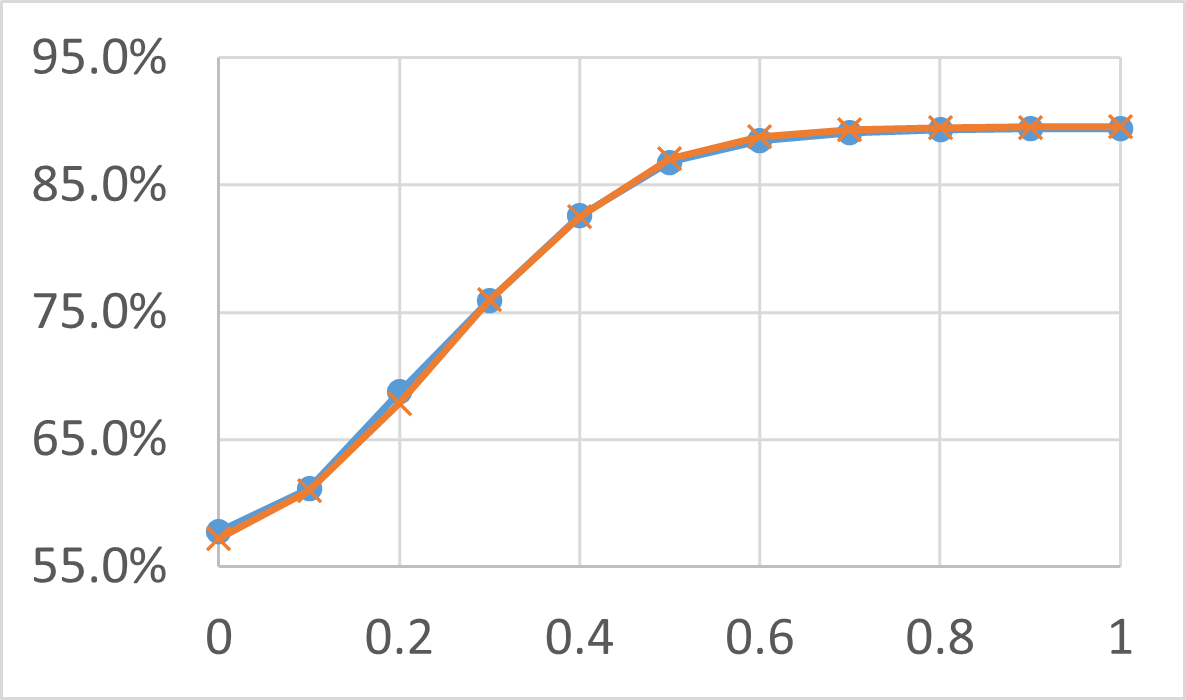}
    \vspace{-.3cm}
    \caption{en-tr}
  \end{subfigure}%
  \begin{subfigure}{.25\textwidth}
    \centering
    \includegraphics[height=2.5cm,width=4.0cm]{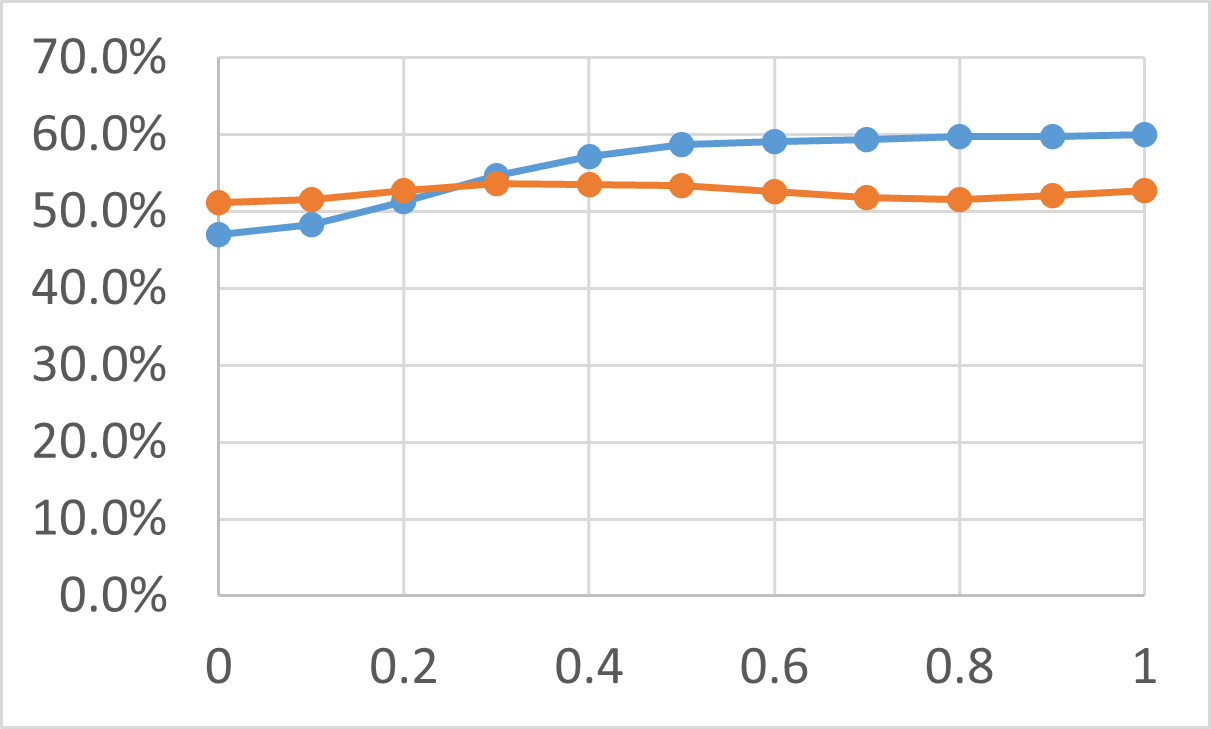}
    \vspace{-.3cm}
    \caption{ha-en}
  \end{subfigure}%
  \begin{subfigure}{.25\textwidth}
    \centering
    \includegraphics[height=2.5cm,width=4.0cm]{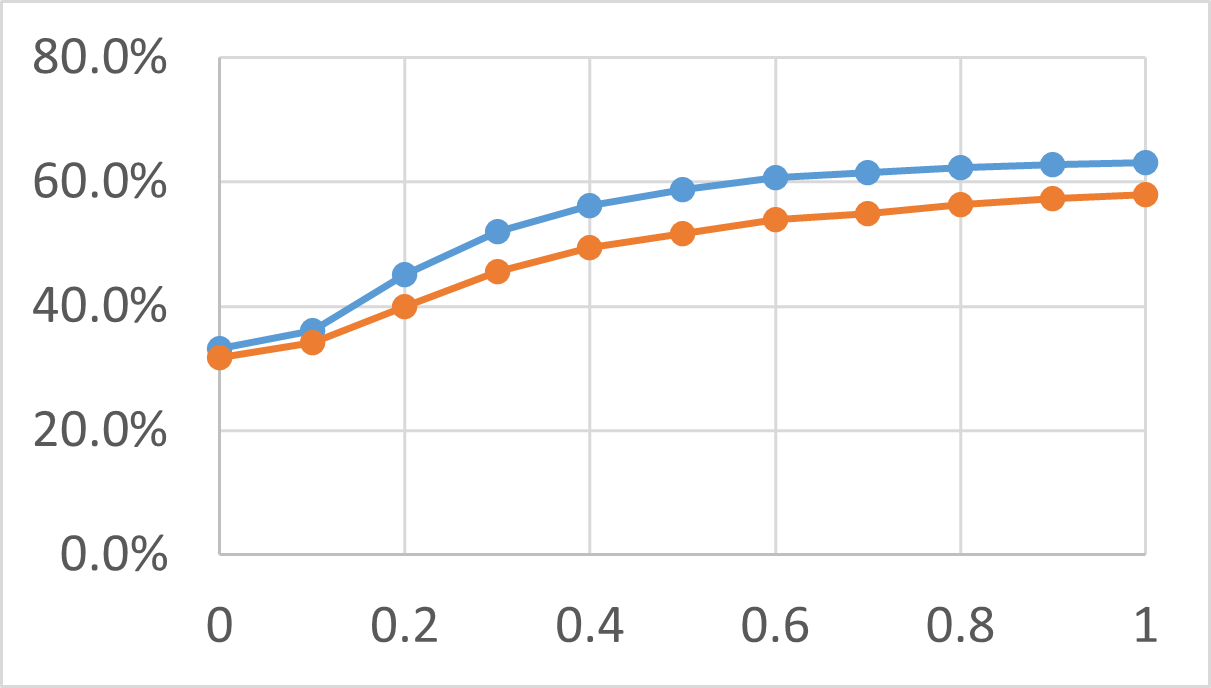}
    \vspace{-.3cm}
    \caption{en-ha}
  \end{subfigure}%

  \caption{Ensembling MT model with 7B parameter LLM. Graphs shows \textsc{Comet-22} vs mixing ratio.}
  \label{fig:7b}
\end{figure*}

\autoref{fig:7b} shows translation quality as we vary the mixing ratio, $\lambda$. 
Note that $p_{\text{\scriptsize ensemble}}$ reduces to the LLM when $\lambda = 0$ and to the MT model when $\lambda = 1$.

For our results in the main section, we selected $\lambda$ on validation set translation quality. 
Here we see that in cases where both models are reasonably strong (de-en, ru-en, and en-ru) the ensembling provides a quality boost.

\begin{figure*}[p]
\centering
  \begin{subfigure}{\textwidth}
    \centering
    \includegraphics[width=.3\linewidth]{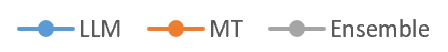}
  \end{subfigure}

  \begin{subfigure}{.25\textwidth}
    \centering
    \includegraphics[height=2.5cm,width=4.0cm]{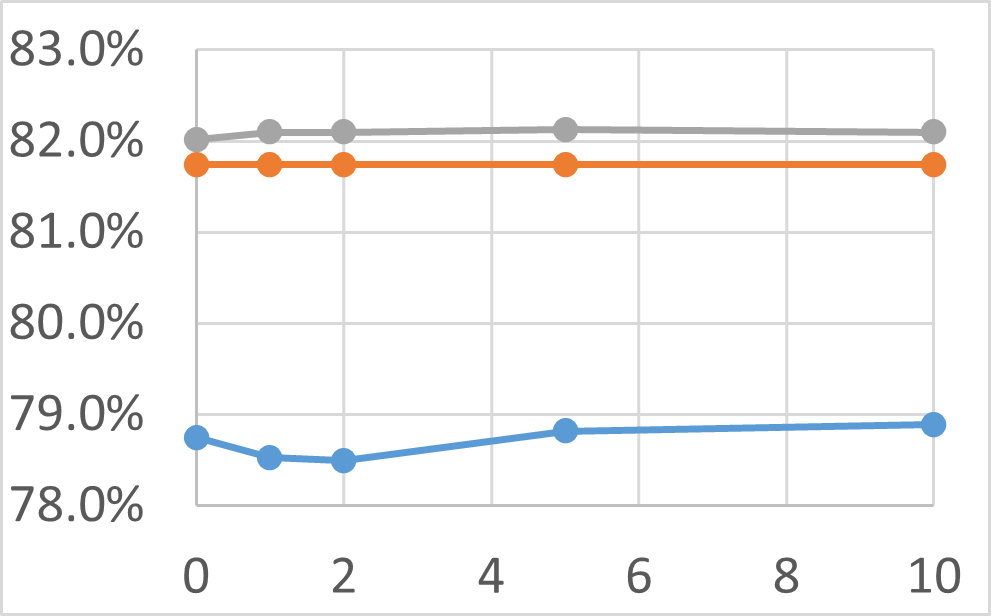}
    \vspace{-.3cm}
    \caption{en-de gender}
  \end{subfigure}%
  \begin{subfigure}{.25\textwidth}
    \centering
    \includegraphics[height=2.5cm,width=4.0cm]{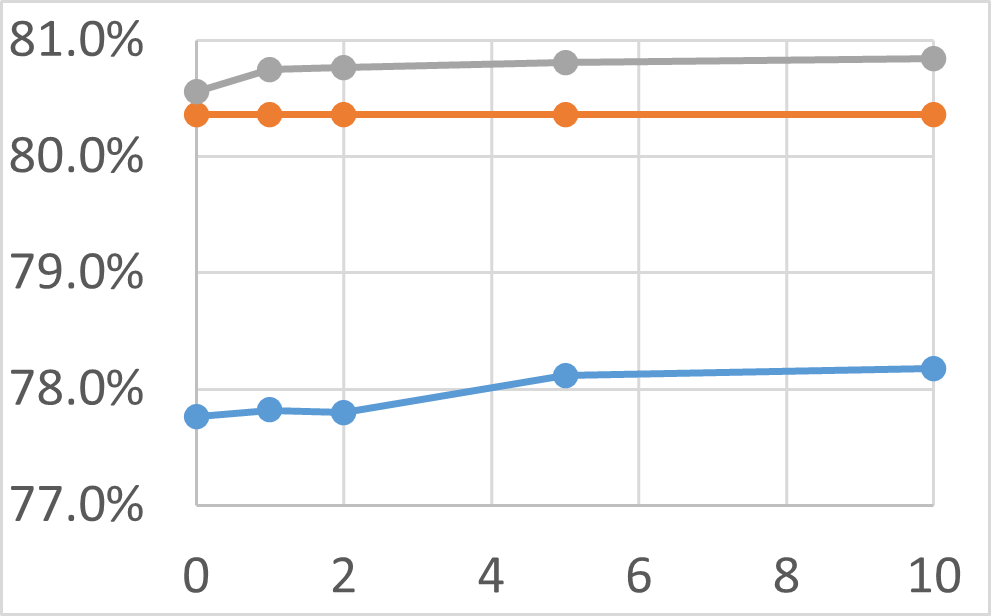}
    \vspace{-.3cm}
    \caption{en-de auxiliary}
  \end{subfigure}%
  \begin{subfigure}{.25\textwidth}
    \centering
    \includegraphics[height=2.5cm,width=4.0cm]{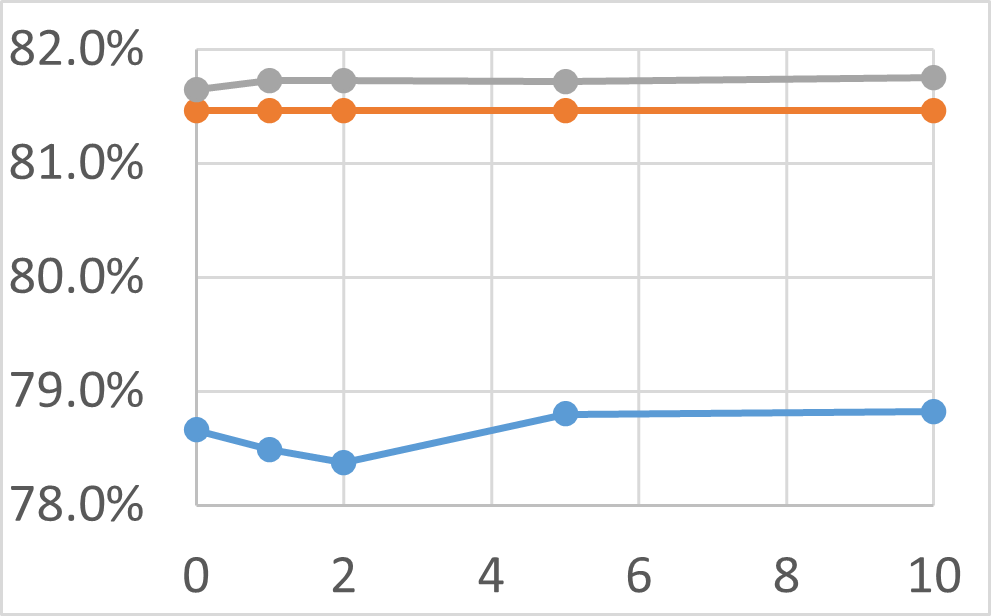}
    \vspace{-.3cm}
    \caption{en-de formality}
  \end{subfigure}%
  
  \begin{subfigure}{.25\textwidth}
    \centering
    \includegraphics[height=2.5cm,width=4.0cm]{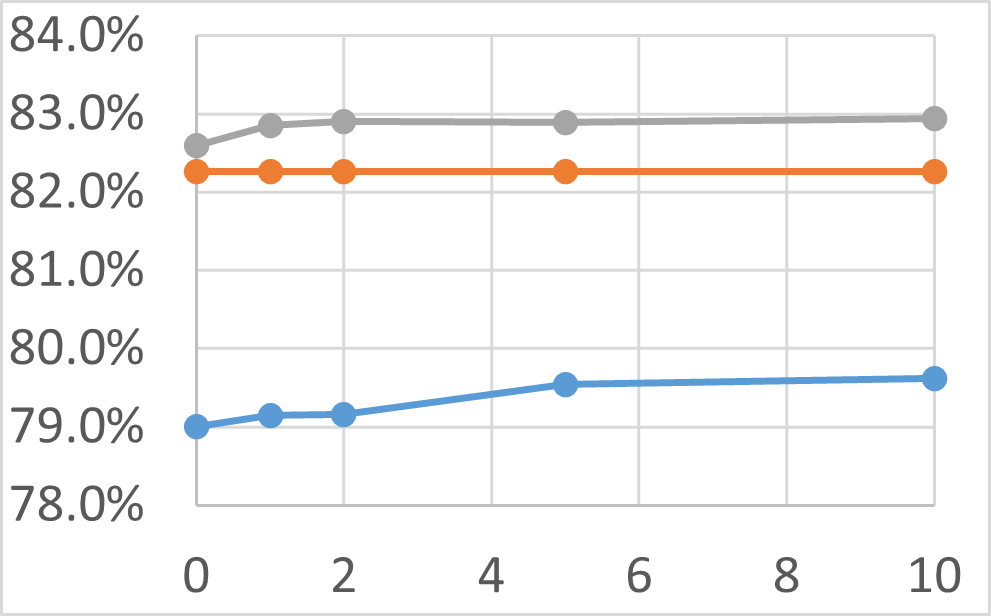}
    \vspace{-.3cm}
    \caption{en-ru gender}
  \end{subfigure}%
  \begin{subfigure}{.25\textwidth}
    \centering
    \includegraphics[height=2.5cm,width=4.0cm]{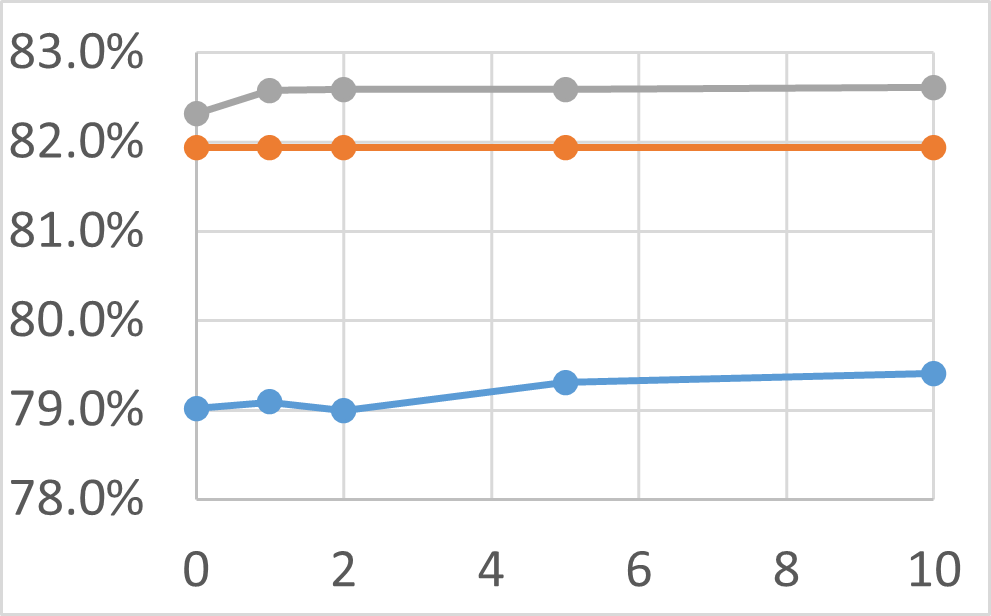}
    \vspace{-.3cm}
    \caption{en-ru inflection}
  \end{subfigure}%
  \begin{subfigure}{.25\textwidth}
    \centering
    \includegraphics[height=2.5cm,width=4.0cm]{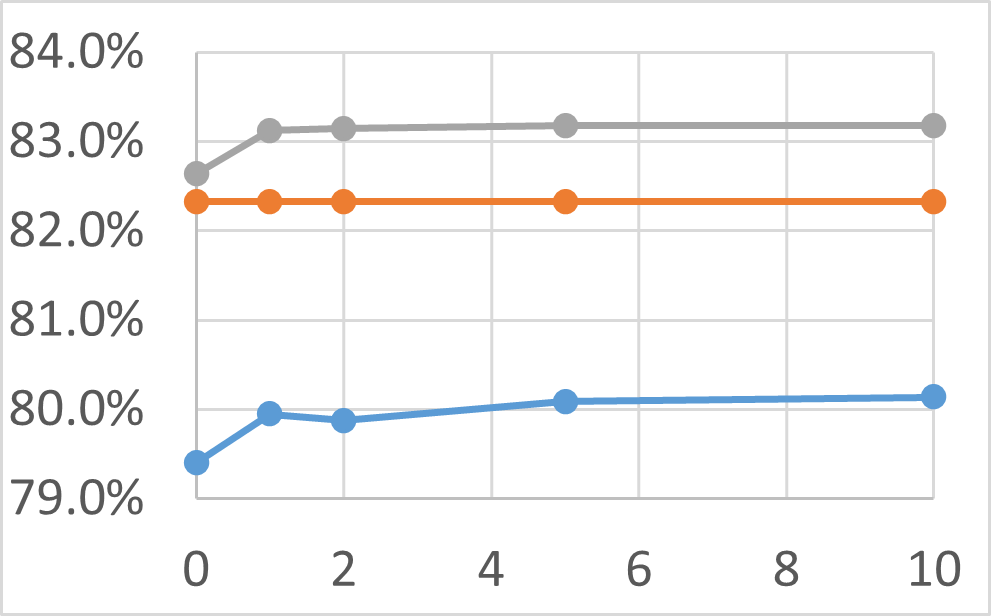}
    \vspace{-.3cm}
    \caption{en-ru auxiliary}
  \end{subfigure}%
  \begin{subfigure}{.25\textwidth}
    \centering
    \includegraphics[height=2.5cm,width=4.0cm]{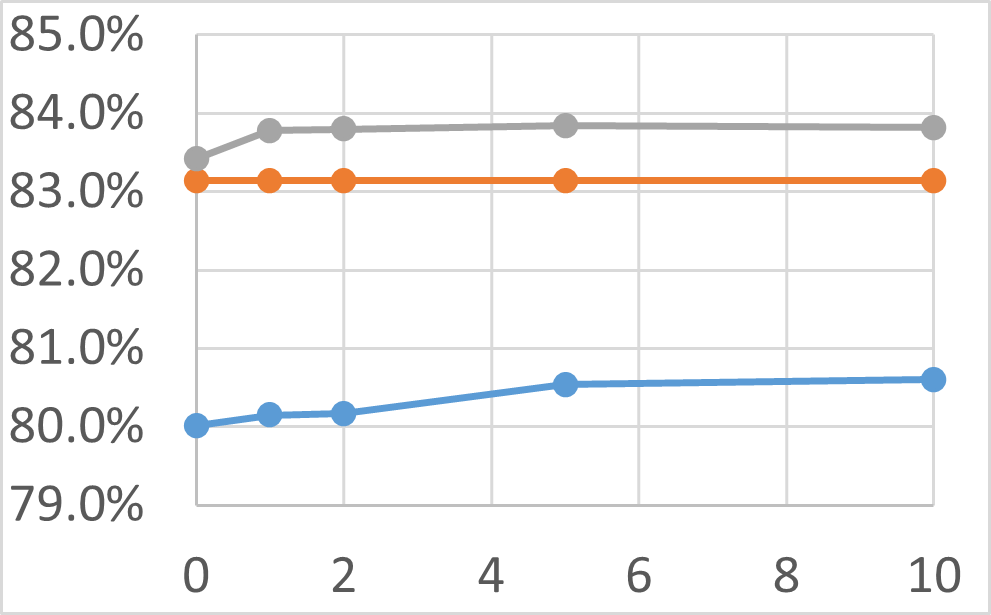}
    \vspace{-.3cm}
    \caption{en-ru formality}
  \end{subfigure}%

    \caption{\textsc{Comet-22} on the data in CTXPro.}
  \label{fig:CTXPro.comet}
\end{figure*}

\subsection{COMET-22 CTXPro}
\label{sec:app_CTXPro.comet}
\autoref{fig:CTXPro.comet} shows the \textsc{Comet-22} scores corresponding to the document translation accuracy show in \autoref{tab:CTXPro.accuracy}. The ensemble is
always best on this data, then the MT, and then the LLM.

\end{document}